\documentclass[letterpaper, 10 pt, conference]{ieeeconf}

\IEEEoverridecommandlockouts                              % This command is only needed if 
\title{\LARGE \bf
Neural Autonomous Navigation with Riemannian Motion Policy
}

\author{Xiangyun Meng, Nathan Ratliff, Yu Xiang and Dieter Fox
\thanks{Xiangyun Meng is with Paul G. Allen School of Computer Science \& Engineering, University of Washington, Seattle, WA, USA {\tt\small xiangyun@cs.washington.edu}}
\thanks{Dieter Fox is with NVIDIA, USA {\tt\small dieterf@nvidia.com} and Paul G. Allen School of Computer Science \& Engineering, University of Washington, Seattle, WA, USA {\tt\small fox@cs.washington.edu}}
\thanks{Nathan Ratliff, Yu Xiang are with NVIDIA, USA {\tt\small \{ nratliff, yux\}@nvidia.com }}
}

\usepackage{xcolor}
\usepackage{amsmath}
\usepackage{amssymb}
\usepackage{siunitx}
\usepackage{graphicx}
\usepackage{adjustbox}
\usepackage{float}
\usepackage[labelformat=simple,font=footnotesize]{subcaption}
\usepackage[font=small]{caption}
\usepackage{booktabs}

\begin{document}

\newcommand{\todo}[1]{\textcolor{red}{\textbf{#1}}}
\newcommand{\mb}[1]{\mathbf{#1}}
\newcommand{\argmin}{\mathop{\mathrm{argmin}}}

\maketitle
\thispagestyle{empty}
\pagestyle{empty}

%%%%%%%%%%%%%%%%%%%%%%%%%%%%%%%%%%%%%%%%%%%%%%%%%%%%%%%%%%%%%%%%%%%%%%%%%%%%%%%%
\begin{abstract}

%We introduce a novel image-based autonomous navigation framework by leveraging the Riemannian Motion Policy (RMP) and deep learning for vehicular control. A deep neural network is designed to predict control point RMPs of the vehicle from visual images, from which the optimal control commands can be computed analytically. We show that our network trained in Gibson environment can be used for indoor obstacle avoidance and navigation on a real RC car, and our RMP representation generalizes better to unseen environments than predicting local geometry or predicting control commands directly.

% Updated by Nathan
End-to-end learning for autonomous navigation has received substantial attention recently as a promising method for reducing modeling error. However, its data complexity, especially around generalization to unseen environments, is high. We introduce a novel image-based autonomous navigation technique that leverages in policy structure using the Riemannian Motion Policy (RMP) framework for deep learning of vehicular control. We design a deep neural network to predict control point RMPs of the vehicle from visual images, from which the optimal control commands can be computed analytically. We show that our network trained in the Gibson environment can be used for indoor obstacle avoidance and navigation on a real RC car, and our RMP representation generalizes better to unseen environments than predicting local geometry or predicting control commands directly.

\end{abstract}

%%%%%%%%%%%%%%%%%%%%%%%%%%%%%%%%%%%%%%%%%%%%%%%%%%%%%%%%%%%%%%%%%%%%%%%%%%%%%%%%
\section{INTRODUCTION}

Creating autonomous robots capable of navigating in complex environments is an important research topic in robotics. Conventional autonomous robots require expensive sensors such as laser scanners to navigate, which only work in specific operating environments \cite{arkin1990autonomous,nashashibi1993,soloviev2007tight} due to physical and sensory constraints. In contrast, cameras are cheap, lightweight and carry rich geometric and semantic information. As a result, image-based navigation receives increasing attention recently \cite{murray2000using,royer2007monocular,de2016hybrid}. Traditional image-based navigation systems are built upon simultaneous localization and mapping (SLAM) \cite{thrun2007simultaneous} techniques, where images are used to reconstruct a 3D world map, whereby a robot localizes and tracks itself. Although significant progress has been made \cite{henry2010rgb,mur2015orb,whelan2016elasticfusion,wang2017stereo}, these techniques still have trouble with textureless environments or fast-moving cameras.

By leveraging the recent advancements of deep neural networks, there exist two new paradigms for image-based autonomous navigation. The first paradigm applies deep learning to predict local geometry from images \cite{eigen2014depth,liu2016learning,godard2017unsupervised,vijayanarasimhan2017sfm}. The idea is to replace the traditional SLAM pipeline with a neural network in order to handle limitations of existing SLAM systems. Using the predicted geometry, traditional planning and control methods can be used for navigation. However, due to the complex mechanics of robots, the error in predicted geometry may result in unpredictable errors in the control commands, potentially causing catastrophic failures. The second paradigm trains a network to map images to control commands directly \cite{codevilla18, Pan2010AgileOA, Xu2017EndtoEndLO}. However, the lack of modelling the geometry and dynamics makes it difficult to interpret the model and also hinders its generalizability.

% there is a growing interest in image-based obstacle avoidance and navigation. Recently purposed neural network based methods \cite{codevilla18, Pan2010AgileOA, Xu2017EndtoEndLO} have shown impressive driving performance by simply learning the controls from data in an end-to-end fashion.

% While impressive results have been achieved, there remain a few limitations that are crucial to precise and explainable vehicle control. First, existing works mostly do not model the vehicle's geometry and dynamics. While one can argue that these properties can be learned from data, the amount of data required can be huge and hence expensive to collect \cite{Xu2017EndtoEndLO}. Second, the learned geometry and dynamics are hidden inside the network, rendering it non-transferable to other robots. For example, the policy for driving a sedan and driving a truck are different. Not only is this approach unscalable considering the great variety of vehicles we have today, the network remains a black box where the behavior of the vehicle cannot be intuitively reasoned about.

\begin{figure}
    \centering
    \includegraphics[width=0.82\columnwidth]{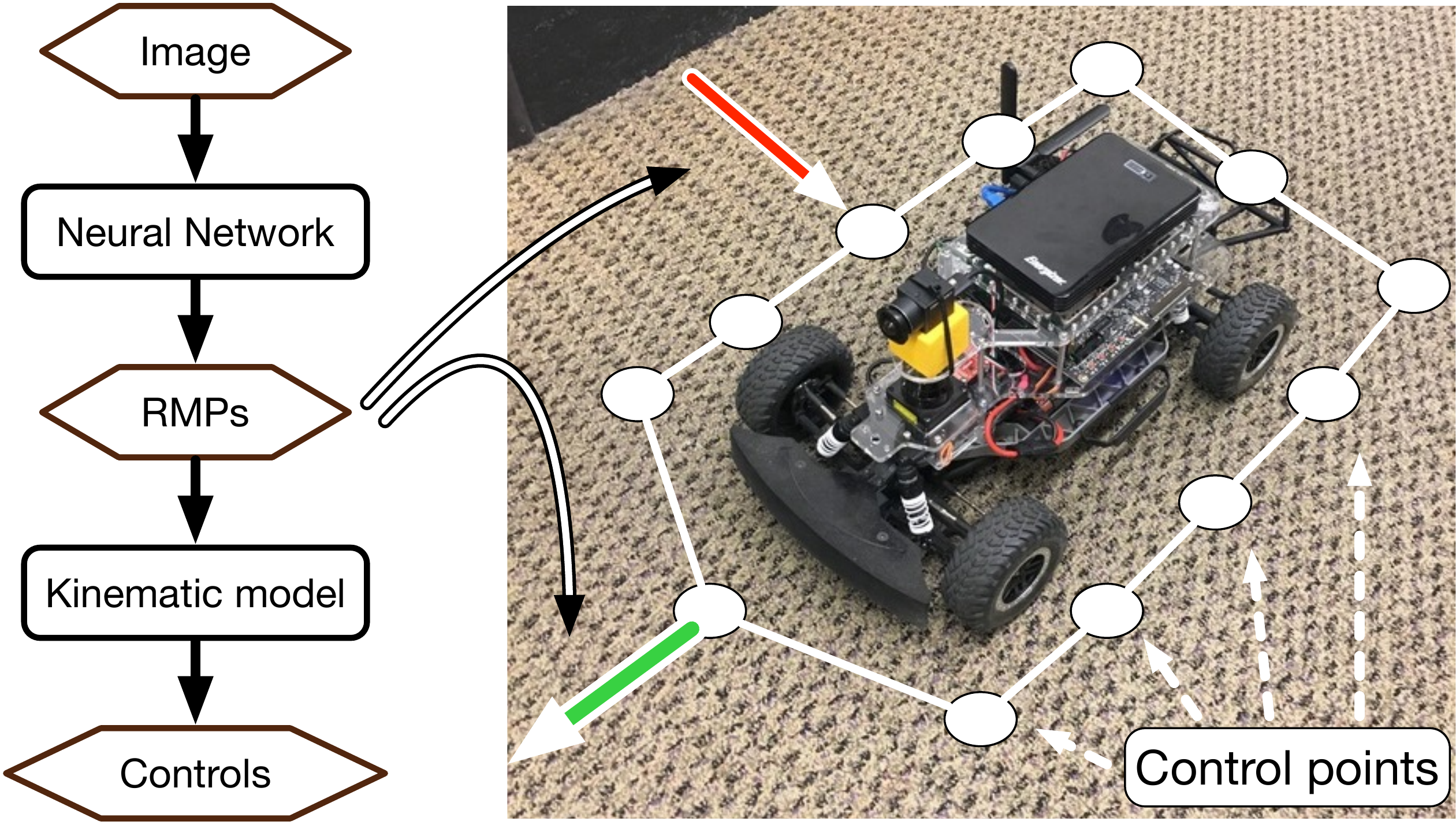}
    \caption{We model the vehicle geometry with a set of control points and the interaction between control points and the environment with RMPs. The interaction can be repulsive (red) or attractive (green). We predict the RMPs from images with a neural network. The RMPs are then combined with the kinematic model of the vehicle to solve for the control commands.}
    \label{fig:intro}
    \vspace{-5mm}
\end{figure}

In this paper, we introduce a new framework for image-based autonomous navigation by leveraging the recently proposed Riemannian Motion Policy (RMP) framework \cite{rmp} and deep learning. RMP is a joint representation of the state of the robot and its environment. It models the interaction between a point on the robot and the environment through an acceleration policy with an associated Riemannian metric. By incorporating the kinematic model, the optimal control commands can be computed analytically. In other words, it unifies the robot geometry, robot dynamics and local obstacles into a single representation. We build this RMP structure into the design of a neural autonomous navigation framework as illustrated in Fig.~\ref{fig:intro}. Our method has the following two main advantages: i) Our neural network trained to predict the RMPs generalizes better to unseen environments compared to networks trained to predict depth or control commands. ii) By examining the predicted RMPs, we are able to reason about the behavior of the vehicle precisely, achieving more explainable neural vehicle control.

% In this paper we propose a new approach of vision-based vehicle control. The key of our approach is to use the recently proposed Riemannian Motion Policy (RMP) framework \cite{rmp} to unify the vehicle's geometry, dynamics and local obstacles into a single representation that can be predicted by a neural network. The optimal control commands can be computed mathematically from this representation. Since the vehicle's geometry and dynamics are specified outside network, our model is more amenable to policy transfer. For example, our learned policy can be transfered easily from a wheeled vehicle to a turtlebot without any modification, whereas the end-to-end approach is not able to do so due to wheeled vehicle being non-holonomic. Moreover, by examining the predicted RMP at each control point on the vehicle, we are able to reason about the vehicle's behavior precisely, achieving more explainable neural vehicle control. \todo{Probably should also mention our local policy can produce complex behaviors not demonstrated by existing neural models.}

We train our RMP controller in the Gibson environment \cite{xia2018gibson}, and evaluate our approach for indoor navigation on a RC car both in simulation and on a real hallway. Our test environments are real floorplans with presence of diverse obstacles and tight spaces, which requires precise maneuver with low tolerance for errors. In particular, the non-holonomic constraint of the vehicle makes steering in tight spaces challenging. We show that the RMP framework solves this problem elegantly and our model generalizes better than predicting control commands or predicting local geometry.

\section{RELATED WORK}

\textbf{Vision-based vehicle control and navigation}. There has been persistent effort since \cite{pomerleau1989alvinn} to map visual input to driving commands with a neural network. Thanks to the advancement of deep learning, recent works have come up with neural policies that achieved impressive driving performance. Most existing works \cite{codevilla18, Pan2010AgileOA, Xu2017EndtoEndLO} adopt a supervised approach to map images to control commands through imitation learning. The expert can be a human operator or a controller learned in advance using non-visual sensors. From this perspective, our approach is also supervised and the expert is a RMP controller. However, our controller differs from most existing works in that our neural network predicts a representation that models the vehicle geometry and dynamics explicitly. % In contrast, mapping images directly to control commands makes it difficult to reason about the network's prediction and could also have generalization issue.

\cite{Mueller2018DrivingPT} presents an approach that decouples perception and control by predicting a waypoint from a segmented image, whereby a PID controller is used to control the vehicle. However, it does not explicitly handle obstacles. Our work takes a step further by predicting RMPs from images. We demonstrate that our RMP representation is a principled approach for local obstacle avoidance and can potentially be used in conjunction with a waypoint predictor to achieve more robust visual autonomous navigation.

Another line of research adopts reinforcement learning (RL) \cite{Kahn2017SelfsupervisedDR, Williams17InfoTheoreticRL, sadeghi2016cad2rl} for self-supervised training. However, the behavior of reinforcement learning depends on the value function (e.g., the probability of collision), which only indirectly defines the behavior of the vehicle. This makes precise control and reasoning of a vehicle difficult. Existing works with RL on driving assume either fixed speed or routes which is impractical for real-world applications.

% Vision-based mapping and planning has also drawn increasing interests. \cite{gupta2017cognitive} presents an end-to-end approach for visual navigation and mapping using a differentiable mapper and planner, but their robot modelling is overly simplified by assuming a point agent with discrete actions. In contrast, our work models the geometric shape and dynamics of a robot. We only require a coarse guidance to the goal, and relies on learned representation for generating smooth control commands for local obstacle avoidance.

% Generalization has always been an issue when transferring models learned from simulation into real world. Since the generalization problem is due to the domain gap between the simulated environment and real environment, domain adaption \todo{cite} becomes a popular technique. Recent works \cite{xia2018gibson} also propose realistic simulation environments to reduce the domain gap, but from our experiments training only on realistic simulation environments does not produce a satisfactory policy. However, finetuning the model with a few minutes of collected data improves the agent's behavior significantly, indicating that bootstrapping from simulation followed by finetuning can be a useful technique for training high-performing neural policies.

\textbf{Model-based robot control}.
Model-based robot control assumes full observation of the environment. The complete knowledge of the geometric structure allows optimal local obstacle avoidance \cite{rmp} and advanced planning algorithms to perform long-range trajectory optimization \cite{riemo, Schulman-RSS-13}. However, the assumption that the geometric structure of the environment is known is unrealistic for a moving vehicle with a monocular RGB camera.

Our approach marries vision-based vehicle control with model-based robot control through the Riemannian metric representation of the robot and its environment. The vision system outputs a representation that can be used in the classical robot control framework. To this end, these two regimes are able to complement each other to overcome their weaknesses.

\section{RMP MODELLING}

In this section, we provide a brief introduction of RMP and its application to vehicle control. See \cite{rmp} for a theoretical introduction in a multi-joint robotic arm setting.

\begin{figure*}[t]
\begin{minipage}[b]{0.33\textwidth}
 \centering
  \includegraphics[width=0.95\columnwidth]{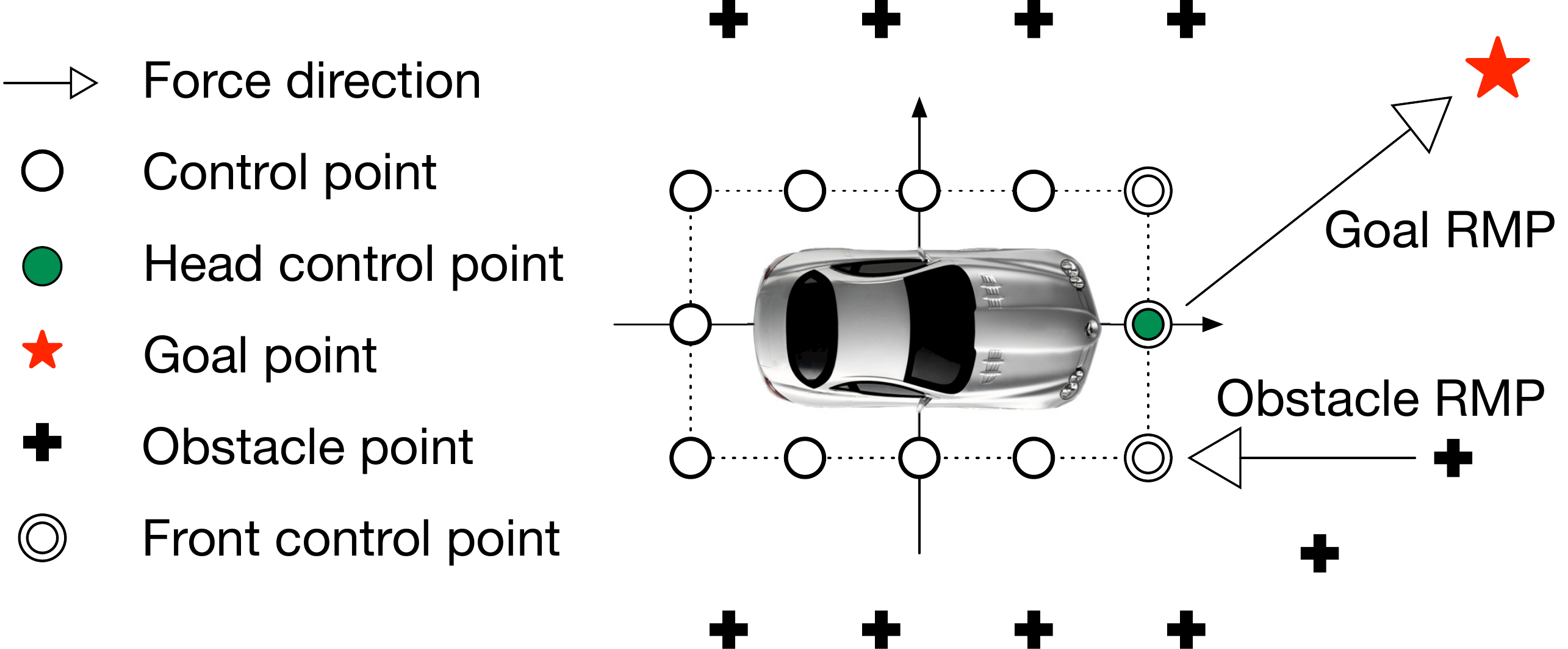}
  \caption{We model the vehicle geometry with control points and its interaction with the environment with RMPs.}
  \label{fig:car_geometry}
\end{minipage}\hfill
\begin{minipage}[b]{0.33\textwidth}
 \centering
  \includegraphics[width=0.95\columnwidth]{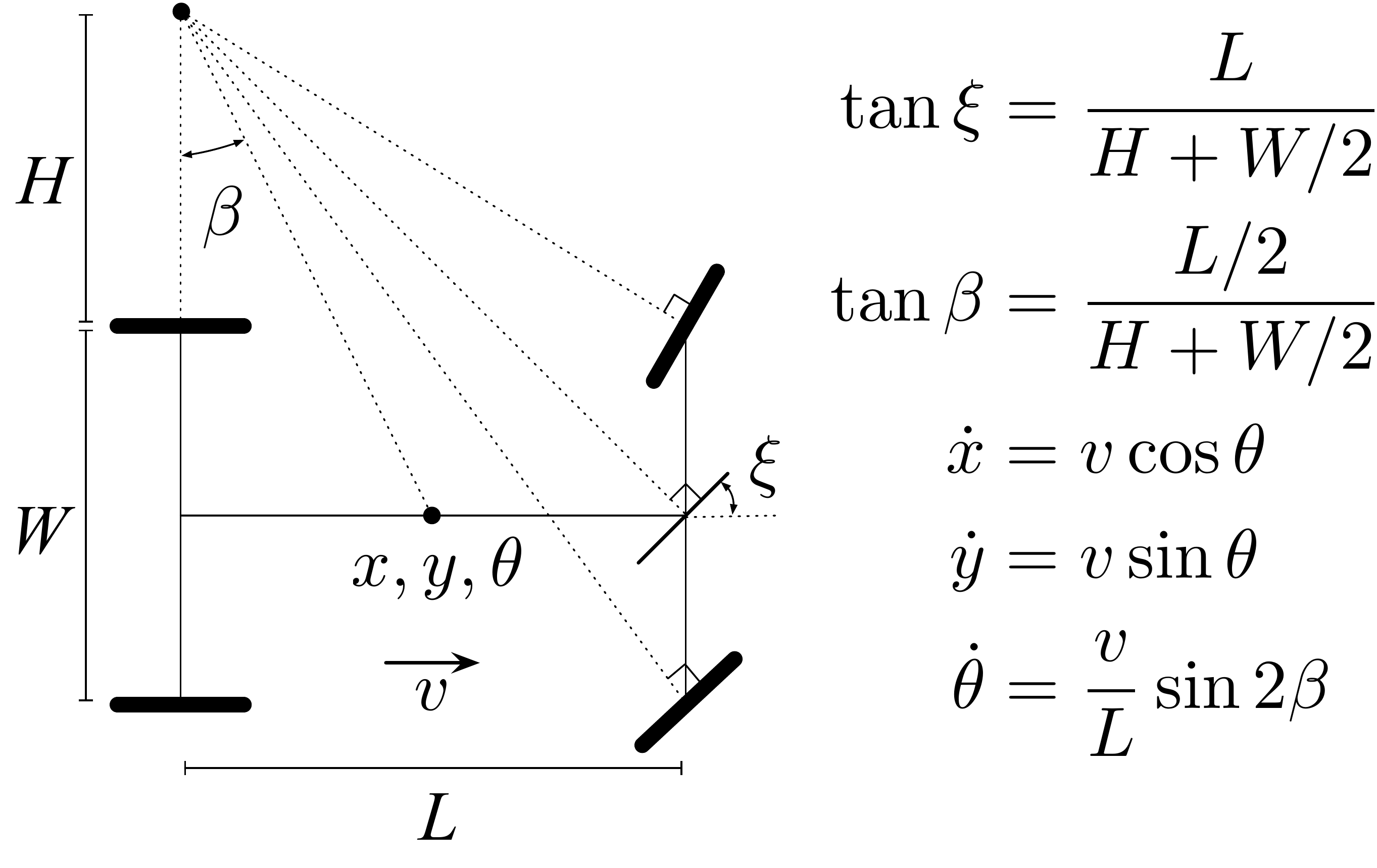}
  \caption{Kinematic model. $v$ is the forward velocity and $\xi$ is the steering angle.}
  \label{fig:kinematic_model}
\end{minipage}\hfill
\begin{minipage}[b]{0.3\textwidth}
 \centering
  \includegraphics[width=0.8\columnwidth]{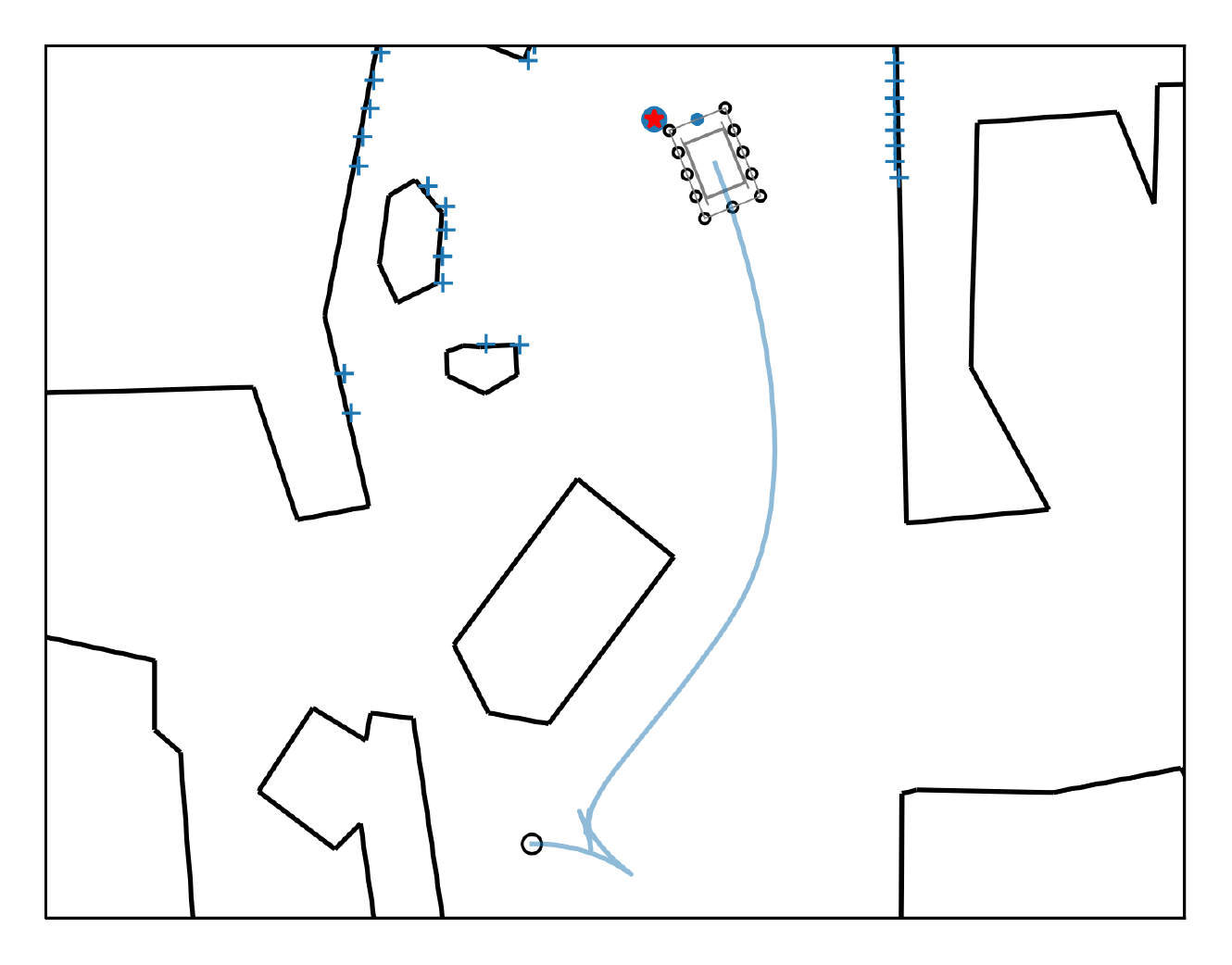}
  \caption{A trajectory generated by our RMP controller.}
  \label{fig:sample_run}
\end{minipage}
\vspace{-4mm}
\end{figure*}

\subsection{A Brief Introduction of RMP}

Consider a point agent $\mathbf{x}$ in $\mathbb{R}^n$ (usually $n=2$ for a ground vehicle). Denote the position of the agent at time $t$ as $\mb{x}(t)$. A motion policy is a mapping $\mb{f}: \mb{x}(t),\dot{\mb{x}}(t) \mapsto \ddot{\mb{x}}(t)$ that maps position and velocity of the agent to an acceleration. The agent applies this acceleration for a small time step to reach a new state $\mb{x}(t+1), \dot{\mb{x}}(t+1)$, whereby a trajectory can be generated through forward integration.

In autonomous navigation tasks, we would like to have a motion policy that guides the agent to its destination $\mb{g} \in \mathbb{R}^n$ while avoiding obstacles along the way. To achieve this goal, we model obstacles as a set of points $\mb{o}_1, ..., \mb{o}_m \in \mathbb{R}^n$, which are directly measurable using commodity sensors such as a laser scanner or a depth camera. Then the problem becomes designing a motion policy that keeps the agent away from the obstacles $\mb{o}_1, .., \mb{o}_m$ while moving towards the goal $\mb{g}$.

An intuitive way to design this motion policy is to divide it into a set of policies that model the interaction between the agent and each goal or obstacle point. For example, for each obstacle, we assign a policy that produces a repelling acceleration towards the agent, whereas for the goal point the policy produces an attractive acceleration. 
%These accelerations are usually functions of relative distances between the agent and the obstacles or goal points. 
The optimal motion policy can be solved in a least-squares manner:
\begin{equation}
    \mb{f}(\mb{x}, \mb{\dot{x}}) = \argmin_{\mb{\ddot{x}}} \sum_i \frac{1}{2} || \mb{f}_i(\mb{x}, \mb{\dot{x}}) - \mb{\ddot{x}} ||^2,
    \label{eq:naive}
\end{equation}
where $\mb{f}_i$ denotes the acceleration of the $i$th obstacle or goal.

Eq.~\eqref{eq:naive} may result in undesirable behaviors. For example, it does not take the error direction into  account. This can make a huge difference when an agent moves in the direction parallel to the obstacle surface compared to moving towards the obstacle surface. A principled solution is to assign a Riemannian metric that stretches the local space such that the cost of moving in one direction is different from another. The magnitude of a cost vector $\mb{v}$ w.r.t a Riemannian metric $\mb{A}$ is defined as $||\mb{v}||_{\mb{A}}^2=\mb{v}^\top\mb{A}\mb{v}$. Hence the policy becomes
\begin{align}
    \mb{f}(\mb{x}, \mb{\dot{x}}) = \argmin_{\mb{\ddot{x}}} \sum_i \frac{1}{2} || \mb{f}_i - \mb{\ddot{x}} ||^2_{\mb{A}_{i}},
    \label{eq:with_metric}
\end{align}
where $\mb{A}_{i}$ is the Riemannian metric associated with the policy $\mb{f}_i$. We define a Riemannian Motion Policy as a motion policy $\mb{f}$ associated with a Riemannian metric $\mb{A}$.

Eq.~\eqref{eq:with_metric} only considers a point agent, but real robots have non-negligible shapes (e.g., a vehicle) and complex mechanics (e.g., a robotic arm). Considering these additional factors, we model the agent as a set of \emph{control points} $\mb{x}_1, \mb{x}_2, ...$ with corresponding forward kinematic functions (also known as a task map) $\phi_i$ linked to a configuration space $\mb{q}$ such that $\mb{x}_i = \phi_i(\mb{q})$. To compute the optimal acceleration in the configuration space, we incorporate the Jacobian of the task map $\mb{J}_\phi=\frac{\partial \phi}{\partial \mb{q}}$ into Eq.~\eqref{eq:with_metric} according to \cite{rmp}:
\begin{align}
    \mb{f}(\mb{x}, \mb{\dot{x}}) = \argmin_{\mb{\ddot{q}}} \sum_i \frac{1}{2} || \mb{f}_i - \mb{J}_{\phi_i} \mb{\ddot{q}} ||^2_{\mb{A}_{i}},
    \label{eq:with_jacobian}
\end{align}
with a slight abuse of notation of $\mb{J}_{\phi_i}$ denoting the Jacobian of a control point involved in motion policy $i$. Eq.~\eqref{eq:with_jacobian} can be solved analytically:
\begin{align}
    \mb{f}(\mb{x}, \mb{\dot{x}}) = \mb{\ddot{q}}^* = (\sum_i \mb{J}_i^\top \mb{A}_i \mb{J}_i)^+(\sum_i \mb{J}_i^\top \mb{A}_i \mb{f}_i),
    \label{eq:solution}
\end{align}
where $^+$ denotes pseudoinverse.

To concretize, consider a turtlebot with local boundary coordinates $\mb{x}_1', \mb{x}_2',...$. The configuration space $\mb{q}$ is its global position and orientation $[x, y, \theta]^\top$, and $\mb{x}_i=\phi_i(\mb{q})$ is the global position of each control point $\mb{R}(\theta) \mb{x}_i' + [x, y]^\top$, where $\mb{R}(\cdot)$ denotes a rotation matrix.

\subsection{RMP Controller for an Ackermann Steering Vehicle}
\label{sec:model}

\subsubsection{RMP Controller Design}
\label{sec:rmp_controller}
Almost all automobiles and RC cars can be modeled as Ackermann steering vehicles. We model the geometry of a  vehicle by defining 12 control points distributed on a rectangular bounding box around the body as in Fig.~\ref{fig:car_geometry}. There are two types of interactions between a control point and a world point:

\textbf{Goal RMP.} The goal point $\mb{g}$ pulls a control point towards it. The acceleration exerted by a goal point is defined as:
\[
\alpha \frac{\mb{g} - \mb{x}}{ ||\mb{g} - \mb{x} + \epsilon||_2} - \beta \mb{\dot{x}}
\]
where $\alpha, \beta$ are gain and damping factors respectively, and $\epsilon$ is a small value to prevent dividing by zero. This expression applies the maximum acceleration when the vehicle is still, and gradually decreases acceleration until the vehicle reaches its maximum speed $\alpha/\beta$. The design of the goal RMP shows an advantage of the RMP framework: the behavior of the vehicle, such as its acceleration curve, can be precisely specified. We use the identity metric $\mb{I}$ for the goal RMP. Since we want the vehicle to point towards the target due to the limited field of view of sensors, we apply the goal RMP only on the three front control points. This will induce a steering force to align the vehicle's heading to the goal.

\textbf{Obstacle RMP.}
An obstacle point pushes a control point away from it. The acceleration exerted by an obstacle $\mb{o}$ to a control point $\mb{x}$ is defined as
\[
\mb{f}_o = -\cfrac{\alpha}{||\mb{o} - \mb{x}||_2} (\beta \mb{u} \cdot \mb{\dot{x}} + \gamma) \mb{u}
\]
where $\alpha, \beta, \gamma$ are gain, damping and offset respectively, and $\mb{u} = (\mb{o} - \mb{x}) / ||\mb{o} - \mb{x}||_2$ which is the unit vector denoting the direction of the obstacle. Intuitively, the force increases when the obstacle is getting closer and when the velocity of the vehicle is pointing towards the obstacle. The metric for an obstacle is defined as
\[
w(||\mb{o} - \mb{x}||_2) \mb{f}_o \mb{f}^\top_o,
\]
where $w(\cdot)$ is a weighting function of the distance to the obstacle that is usually monotonically decreasing. $\mb{f}_o \mb{f}^\top_o$ defines a metric that penalizes directions towards the obstacle and assign zero cost to its null space (i.e., when the agent moves parallel to the obstacle surface).

Combining the goal RMP and the obstacle RMPs yields a usable but unsatisfactory policy. The vehicle tends to wiggle as it moves forward. To stabilize the vehicle, we scale up accelerations of the front left and front right control point. In addition, we add a RMP to the head control point to apply a torque that dampens excessive angular oscillation.

\subsubsection{Incorporating the Kinematic Model}

An Ackermann steering vehicle is non-holonomic, i.e., it cannot move sideways (assuming no drift). Using $\mb{q}=[x, y, \theta]$ will thus not satisfy this constraint and render the policy non-applicable in some situations. To solve this problem, we incorporate the kinematic model of the vehicle into our policy, shown in  Fig.~\ref{fig:kinematic_model} \cite{snider2009automatic}. We parameterize $\ddot{\mb{q}}$ with $[\dot{v},\dot{\xi}]$, i.e., the forward acceleration and steering velocity. We can thus derive the following relations between $[\ddot{x}, \ddot{y}, \ddot{\theta}]$ and $[\dot{v},\dot{\xi}]$:
% \begin{align}
% \ddot{\theta} &= \cfrac{\dot{v}}{L} \sin{2\beta} + \cfrac{v}{L}\cos{2\beta}\cfrac{4}{3\cos^2{\xi}}\dot{\xi}, \\
% \ddot{x} &= \dot{v}\cos{\theta} - \cfrac{v^2\sin{\theta}}{L}\sin{2\beta}, \\
% \ddot{y} &= \dot{v}\sin{\theta} + \cfrac{v^2}{L}\cos{\theta}\sin{2\beta}.
% \label{eq:transform}
% \end{align}
% Writing the above equations in matrix form, we have:
\begin{align*}
\begin{bmatrix}
    \ddot{x}\\
    \ddot{y}\\
    \ddot{\theta}\\
\end{bmatrix}=
\underbrace{
\begin{bmatrix}
\cos{\theta} &0\\
\sin{\theta} &0\\
\frac{\sin{2\beta}}{L} & \frac{4v\cos{2\beta}}{3\cos^2{\xi} +1}\\ 
\end{bmatrix}
}_{\mb{J}}
\begin{bmatrix}
\dot{v}\\
\dot{\xi}\\
\end{bmatrix}+
\begin{bmatrix}
0\\
\frac{v^2}{L}\cos{\theta}\sin{2\beta}\\
0\\
\end{bmatrix}.
\end{align*}
Note that the relationship is not linear due to the additional term. We approximate the relationship by dropping the last term (alternatively we can subsume it into $\mb{f}_i$), allowing us to express Eq.~\eqref{eq:with_jacobian} as
\begin{align}
    \dot{v}^*,\dot{\xi}^* = \argmin_{\dot{v},\dot{\xi}} \sum_i \frac{1}{2} || \mb{f}_i - \mb{J}_{\phi_i} \mb{J} [\dot{v}, \dot{\xi}]^\top  ||^2_{\mb{A}_{i}}.
    \label{eq:rmp_car}
\end{align}
%We found this approximation works well in both simulation and real experiments.
% In principle we should be able to solve the equation exactly \todo{cite}, but we found this approximation works well in both simulation and real experiments. An exact derivation would require excessive space so we leave it as future work \yuxiang{rephrase, not convincing}.

Fig.~\ref{fig:sample_run} shows a simulated trajectory of our RMP controller on a real floorplan. The vehicle starts with facing the wall and the goal is behind an obstacle. No planner is used and the motion of the vehicle is solely based on a simulated $240^\circ$ laser scanner installed on the vehicle. Our RMP controller naturally considers the motion constraints of the vehicle and initially exerts a backward acceleration with a left steering signal, adjusting the heading of the vehicle towards the goal. Once the heading is good enough, it produces a forward acceleration with a right steering signal to curve around the obstacle and eventually reach the goal. This example shows that even our motion policy is based on local observations only, it is able to produce complex behaviors that are useful for real-world navigation. Interestingly, we find it resembles how human drives in this example.

\section{NEURAL RMP}

\label{sec:neural}
The RMP controller described in Sec.\ref{sec:model} requires coordinates of the obstacle points, which cannot be directly measured using a monocular RGB camera. In this section, we present three neural architectures to learn RMP controllers from visual images: two baseline models that predict depth and controls respectively, and our model that predicts RMPs.

\textbf{Predicting depth.} In order to obtain the coordinates of the obstacle points, a straightforward way is to design a neural network that predicts a 1D laser scan (i.e., a depth map) from an input image, from which we apply the RMP controller analytically. However, predicting depth from a single image may not generalize well, as we shall see in Sec~\ref{sec:exp}. Secondly, visual depth estimation can be ill-posed, such as when the vehicle is facing a wall where the image becomes completely featureless. With depth information becoming highly uncertain, the behavior of the vehicle also becomes unpredictable.

\textbf{Predicting controls.} An alternative scheme is to train an end-to-end network \cite{codevilla18, Pan2010AgileOA, Xu2017EndtoEndLO} that directly outputs the control commands. This is appealing due to its ability of learning both geometric and semantic information from visual images. However, this approach is entirely data-driven and thus lacks an interpretation of how the environment affects the behavior of the vehicle. Furthermore, without explicitly modelling the geometry and dynamics of the vehicle, it could also have an adverse effect on the generalizability of model, which we show in our experiments in Sec~\ref{sec:exp}.

\textbf{Predicting RMPs.} To address the limitations of the abovementioned approaches, we propose a new model that predicts RMPs from visual images. RMP has merits from both schemes. The acceleration component $\mb{f}$ in a RMP is the command applied to a control point, whereas the metric component $\mb{A}$ encodes the local geometry of that control point. Moreover, by incorporating the control point Jacobians and the kinematic model, we can solve the optimal control command for the vehicle by combining the contribution of each control point in a geometrically and kinematically consistent manner, potentially having a more interpretable and generalizable model.

\begin{figure}
 \centering
  \includegraphics[width=0.8\columnwidth]{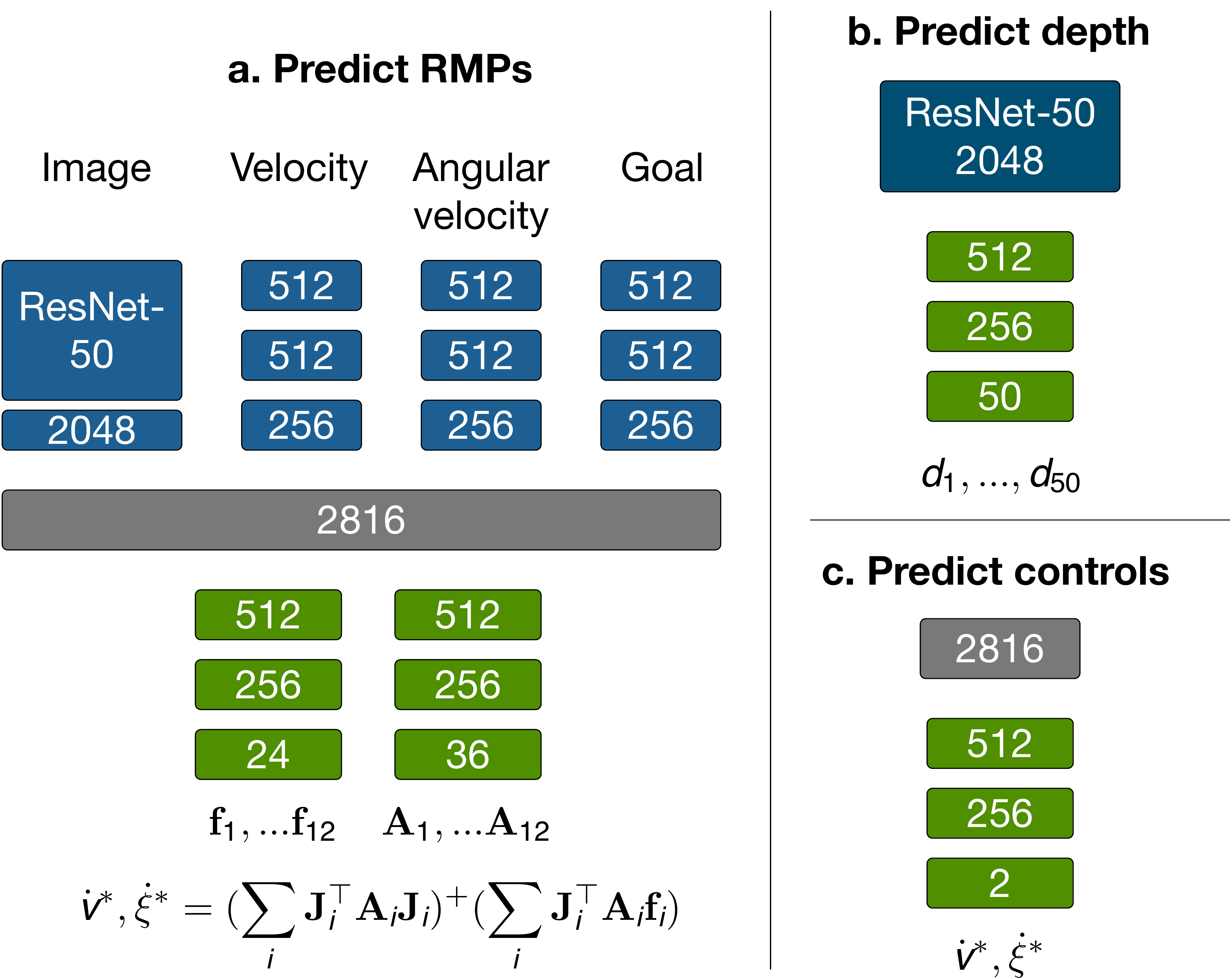}
  \caption{Our RMP network and two baseline models. The predicting controls network reuses the feature extractor of the RMP network.}
  \label{fig:network}
  \vspace{-6mm}
\end{figure}

Fig.~\ref{fig:network}a shows the architecture of our neural RMP model. It comprises multiple feature extractors, a regressor and a solver. The image feature extractor can be any image classification network. We adopted the pretrained ResNet-50 \cite{he2016deep} because it produces good features while not being too heavyweight to run on an embedded computer (e.g., Jetson TX2). Since a RMP may require additional information as inputs, such as velocity or goal point location, we add a feature extractor (multiple fully connected layers) for each type of odometry information. The features are then concatenated and fed into a regressor (multiple fully connected layers) to predict an array of RMPs, i.e., accelerations and metrics. % We do not freeze any part of the network during training, allowing gradients to backpropagate through the entire network.

While we may define multiple RMPs for each control point, the RMP framework allows combining multiple RMPs into a single equivalent RMP \cite{rmp}, hence we only predict one RMP for each control point. For a land vehicle, a RMP consists of a 2-element acceleration $\mb{f}$ and a $2\times 2$ Riemannian metric. Due to the symmetric nature of Riemannian metrics, we only predict three values for each metric, from which we assemble the full matrix. Finally, the solver merely computes the final control command, and is not part of the network.

We also propose two models that predict depth and controls respectively for comparison. The first model (Fig.~\ref{fig:network}b) predicts depth from image features. The second model (Fig.~\ref{fig:network}c) replaces our RMP regressor with a control command regressor. We keep the architecture and computational cost roughly the same for all models so that we can compare which representation is more effective.

\begin{table*}
  \centering
  \footnotesize
  \begin{tabular}{l|*{10}{c}}
    \toprule
    \textbf{Agents}         
    & \multicolumn{2}{c}{\textbf{space8}} 
    & \multicolumn{2}{c}{\textbf{house24}} 
    & \multicolumn{2}{c}{\textbf{house29}}
    & \multicolumn{2}{c}{\textbf{house31}}
    & \multicolumn{2}{c}{\textbf{house57}}\\
    & reached & collision 
    & reached & collision 
    & reached & collision 
    & reached & collision 
    & reached & collision\\
    \hline
    Expert         
    & 97.9\% & 0.4\% 
    & 59.2\% & 3.6\% 
    & 94.5\% & 2.4\% 
    & 97.0\% & 0.9\%
    & 95.0\% & 1.4\%\\
    \hline
    Predicting RMPs 
    & \textbf{88.1}\% & \textbf{7.4}\% 
    & 75.5\% & \textbf{5.2}\%
    & \textbf{93.7}\% & \textbf{1.6}\%
    & \textbf{82.1}\% & \textbf{3.6}\%
    & \textbf{89.5}\% & \textbf{5.9}\%
    \\
    Predicting depth 
    & 85.4\% & 10.3\%
    & \textbf{79.1}\% & 10.1\%
    & 89.0\% & 9.4\%
    & 73.3\% & 22.5\%
    & 78.1\% & 17.8\%
    \\
    Predicting controls 
    & 51.3\% & 19.9\%
    & 48.0\% & 16.3\%
    & 68.5\% & 14.2\%
    & 47.4\% & 25.5\%
    & 56.6\% & 21.9\%
    \\
    \bottomrule
  \end{tabular}
  \caption{Statistics on the five holdout Gibson environments. Note that our neural RMP occasionally outperforms the expert (house24). This is because our neural model is less conservative in avoiding obstacles which leads to higher reach\% in  environments with narrow passages, but at a cost of higher collision\%. The collision events for our expert controller happened mostly when the vehicle at high speed entered a narrow passage from an open space. Additional tuning could reduce such collision events.}
  \label{tab:simulation_comparison}
  \vspace{-2mm}
\end{table*}

\begin{figure*}
    \centering
    \includegraphics[width=0.9\textwidth]{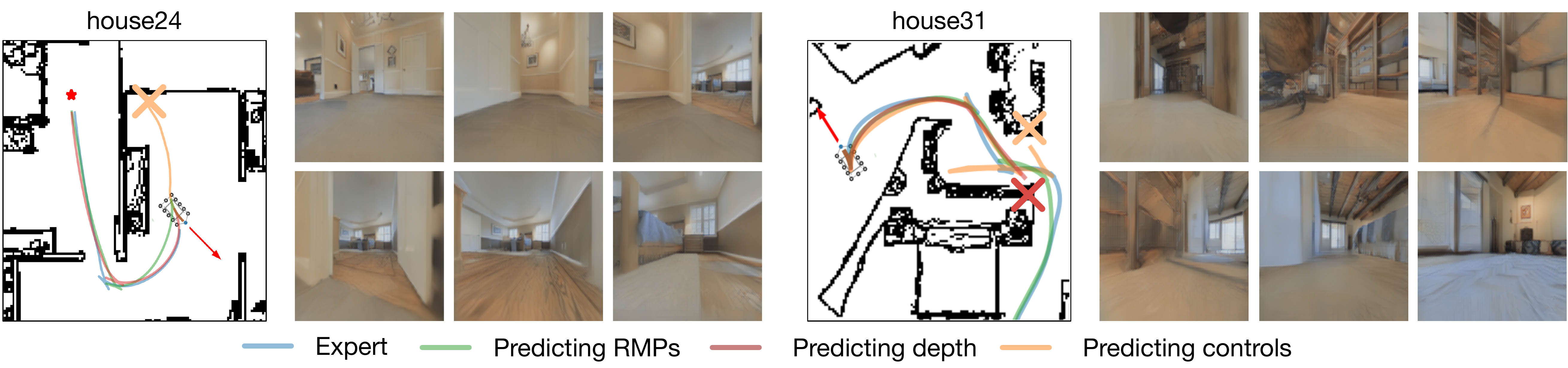}
    \caption{Sampled trajectories in our test sets. Red arrow shows the initial location and heading. Cross shows the collision point.}
    \label{fig:sim_trajs}
    \vspace{-5mm}
\end{figure*}

\section{SYSTEM IMPLEMENTATION}
We designed our RMP controller for a 1/10 RC car \cite{racecar} (Fig.~\ref{fig:intro}). The vehicle has a dimension of approx. $40\si{\cm} \times 25\si{\cm}$, and is equipped with a $240^\circ$ field of view laser scanner. The vehicle takes speed and steering angle as control commands while providing current speed and steering angle as odometry information. The laser scanner is used to localize the vehicle to compute waypoints \cite{fox1999markov}, and also to test our expert RMP controller. Note that we do not require the waypoints to be precise or even visible (see Sec.\ref{sec:real_obstacle_avoidance}), so other localization systems such as Wi-Fi based methods can also be used. We manually tuned the RMP parameters for the RC car. We found that after our expert (RMP using laser scans), works well in simulation, it requires little tuning on the real vehicle, indicating that our simulated expert transfers well to the real vehicle.

The vehicle is also equipped with a fisheye RGB camera and a Jetson TX2 computer to run our neural model. Fisheye images are rectified and cropped to produce images of $120^\circ$ horizontal field of view. % with a resolution of $224\times 224$. 
%We optimized our inference pipeline to achieve a 25 Hz control loop throughput.
Since we are doing closed-loop control, minimizing latency is important. We optimized our model using the TensorRT engine to achieve 50 fps inference time from a $224\times224$ RGB image. Taking other factors into consideration, such as image capture, image rectification and RMP solving,  our end-to-end control loop runs at 25 fps.

\section{EXPERIMENTS}
\label{sec:exp}

We trained our neural model in the Gibson simulation environment \cite{xia2018gibson}. The groundtruth trajectories are generated using the expert RMP controller ( Sec.\ref{sec:model}) with a $240^\circ$ laser scanner and RMP parameters tuned for the real RC car. Our training dataset consists of 60k trajectories sampled from 15 indoor spaces. During training, we randomized velocity (scale $\sim\mathcal{U}(0.0, 2.0)$, rotation $\sim\mathcal{U}(0, 2\pi)$), waypoints (rotation $\sim\mathcal{U}(0, 2\pi)$) and lighting conditions (contrast $\sim\mathcal{U}(0.5, 2.0)$ and brightness $\sim\mathcal{U}(-0.2, 0.2)$). We use the L2 loss to compute gradients. After the first epoch, we apply DAGGER \cite{ross2011reduction} to augment the dataset and gradually increase the ratio of DAGGER samples for subsequent epochs.

% \begin{figure}[t]
%   \centering
%   \includegraphics[width=1.0\columnwidth]{figures/sample_floorplan.pdf}
%   \caption{A sample Gibson environment. The left image shows the floorplan. The right image shows the rendered image by putting the camera at the red dot while pointing in the arrow's direction.}
%   \label{fig:images}
% \end{figure}

% \begin{figure}[!t]
% \centering
% \begin{subfigure}{0.49\columnwidth}
%   \centering
%   \includegraphics[width=0.5\textwidth]{figures/network_depth.eps}
%   \caption{Regressing depth}
%   \label{fig:network_depth}
% \end{subfigure}\begin{subfigure}{0.49\columnwidth}
%   \centering
%   \includegraphics[width=0.5\textwidth]{figures/network_accel.eps}
%   \caption{Regressing controls}
%   \label{fig:network_accel}
% \end{subfigure}
% \caption{Baselines \todo{Explain}}
% \label{fig:baselines}
% \end{figure}

\subsection{Comparing Predicting Depth, Controls and RMPs}

We train one agent for each approach using the same training dataset for the same number of epochs, and test them in 5 holdout Gibson environments. During testing, we randomly sample a starting point and a goal point, and use A* to compute a shortest path. At any time step, the goal point of the agent is the furthest visible point on the shortest path, hence the agent only uses the path as a coarse guidance, and does not have to follow it strictly. In fact, the shortest path is greedy (e.g., very close to obstacles) so that naively following it would cause collision.

\begin{figure*}
  \centering
  \includegraphics[width=0.98\textwidth]{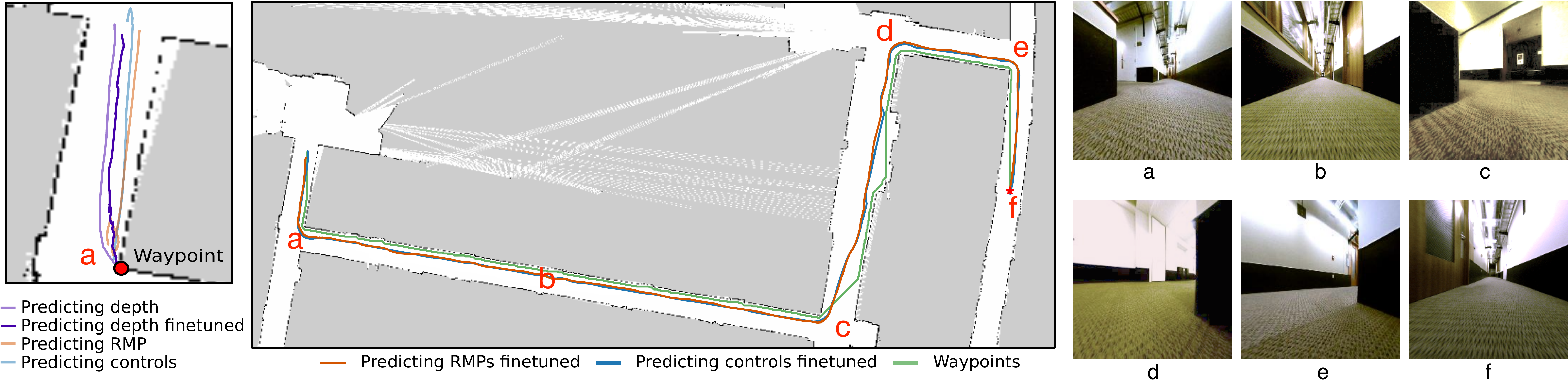}
  \caption{Recorded trajectories in a real hallway. Left: before finetuning all agents hit the walls. Middle: finetuned \emph{Predicting RMPs} and \emph{Predicting controls} agents reached the goal. Right: sample images on the trajectory.}
  \label{fig:hallway_nav}
  \vspace{-3mm}
\end{figure*}

We use two metrics to evaluate the navigation performance: the percentage of trajectories where an agent reaches the goal (reached\%), and the percentage of trajectories where collision occurs (collision\%). We stop an agent once collision happens, hence we have $\text{reached\%} + \text{collision\%} + \text{stuck\%} = 100\%$.
We collected over 200 trajectories for each holdout environment and present the results in Table~\ref{tab:simulation_comparison}. The performance of our RMP agent is the closest to our expert, with comparable high reached\% and low collision\%. Both \emph{Predicting depth} and \emph{Predicting controls} have much higher collision rate with lower reached\%, with predicting controls being more likely to get stuck. This shows our RMP agent generalizes much better than predicting controls due to its explicit modelling of the vehicle geometry and dynamics. Compared to predicting depth, the RMP representation is more concise and less noisy, and thus it is more robust when a robot operates in tight spaces, where small measurement errors in the geometry would cause failures.

% \begin{figure}[H]
%   \centering
%   \includegraphics[width=1.0\columnwidth]{figures/sample_sim_trajs.pdf}
%   \caption{Two sample trajectories from our test sets. The red arrow indicates the initial location and heading of all agents.}
%   \label{fig:sample_sim_trajs}
% \end{figure}

Fig.~\ref{fig:sim_trajs} shows two sample test trajectories. These two trajectories are challenging because they require sharp turns that cannot be completed without backing the vehicle due to its steering limit. Also the tight spaces have low tolerance for depth measurement error. \emph{Predicting controls} backs too much in both environments and ends up hitting the walls. The \emph{Predicting depth} agent failed the last turn in house31. In comparison, \emph{Predicting RMPs} succeeded in both cases.

\begin{figure}
    \centering
    \includegraphics[width=0.95\columnwidth]{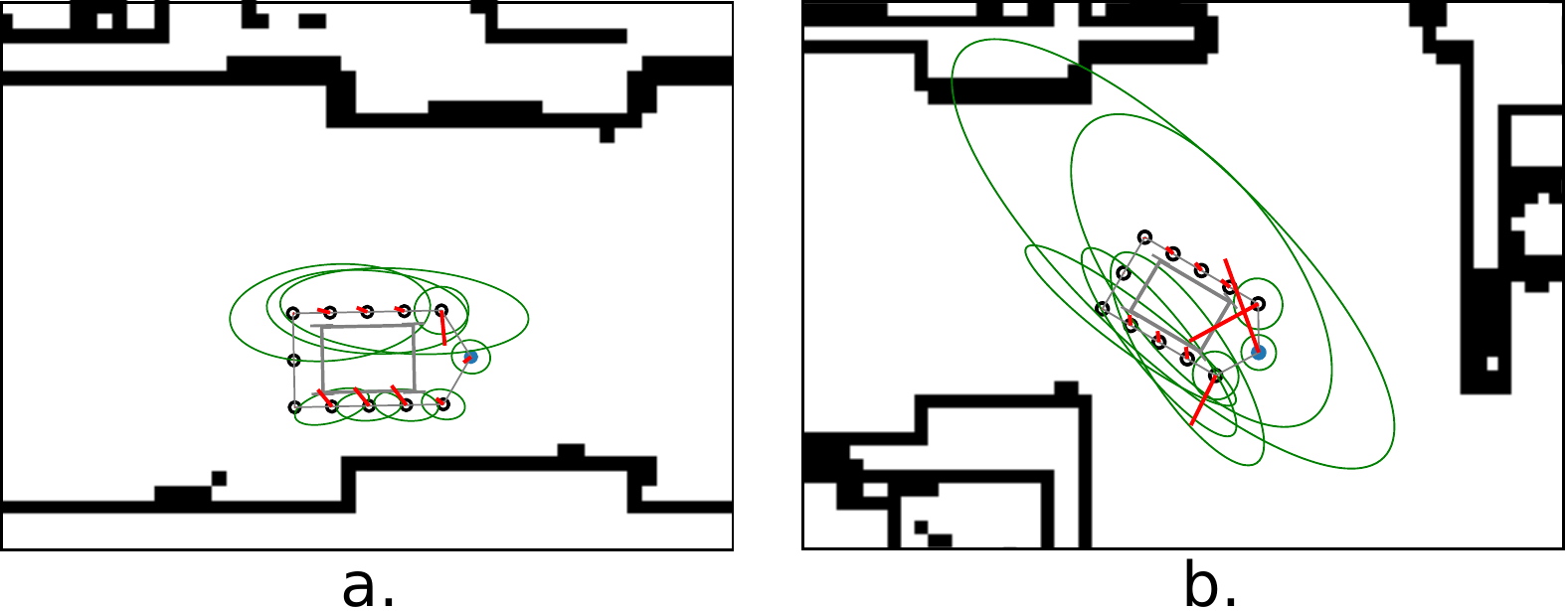}
    \caption{Illustration of predicted RMPs. Red lines are the predicted control point accelerations. Green circles are isocontours of the predicted control point metrics.}
    \label{fig:rmp_vis}
    \vspace{-2mm}
\end{figure}

\subsection{Visualization of the Learned RMPs}
\label{sec:visualization}

Fig.~\ref{fig:rmp_vis} visualizes the predicted RMPs when the vehicle is a) running in a straight corridor and b) doing a right turn. The learned Riemannian metrics assign high costs to the directions towards obstacles as shown by the elliptical isocontours of the metrics, and lower costs to control points far away from obstacles as shown by their larger contour sizes. Same reasoning also holds for the predicted accelerations. As a result, we are able to intuitively reason about the behavior of the vehicle by examining the predicted RMPs. This also allows us to add additional RMPs to adjust the behavior of the vehicle without retraining the network.

\subsection{Real World Experiments}

% \begin{figure*}
% \adjustbox{valign=b}{
% \begin{minipage}[t!]{0.66\textwidth}
% nd{minipage}}
% \adjustbox{valign=b}{\begin{minipage}[t!]{0.33\textwidth}
%  \includegraphics[width=1.0\textwidth]{figures/real_obstacle_avoidance_3_combined.pdf}
%  \caption{\todo{Explain}}
%   \label{fig:real_obstacle_avoidance}
% \end{minipage}}
% \end{figure*}

\begin{figure}[t]
  \centering
  \includegraphics[width=0.98\columnwidth]{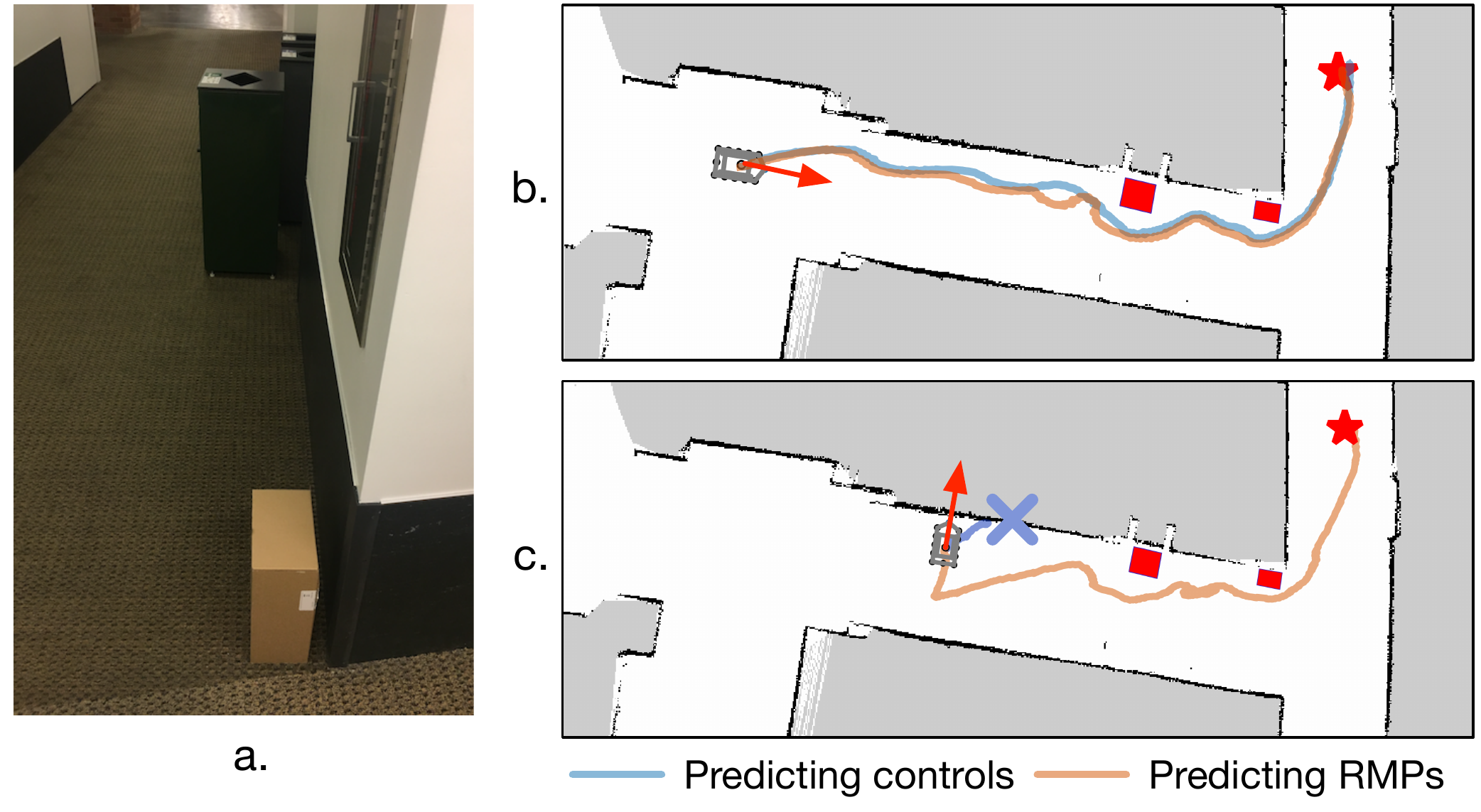}
  \caption{Obstacle avoidance without a planner. Left: placed obstacles. Right: results for two different placements of the vehicle. Red arrow indicates the vehicle heading. Cross shows the collision point.}
  \label{fig:real_obstacle_avoidance}
  \vspace{-6mm}
\end{figure}

\subsubsection{Hallway navigation}
We also evaluated our models trained in simulation in a real hallway to test their generalizability. We perform a similar evaluation as in the Gibson environments by setting a starting point and a goal point and let each agent follow a sequence of waypoints. Fig.~\ref{fig:hallway_nav} shows the recorded trajectories for the three neural agents.

We find that there exists a domain gap between the rendered images in the Gibson environments (Fig.~\ref{fig:sim_trajs}) and the real images taken from the hallway (Fig.~\ref{fig:hallway_nav}). All three agents failed to pass the first left turn because they ran straightly towards the first waypoint which sits at the corner of the wall. While Gibson provides a \emph{Goggles} mechanism \cite{xia2018gibson} to reduce the domain gap, we find its high computational cost introduces too much delay to our control loop. To mitigate this issue, we manually drove the vehicle in the hallway for a few minutes to collect about 13k images with associated laser scans, and annotated the data using our expert RMP controller. We finetuned the three models for a few epochs, which took about 20 minutes each. After finetuning, our \emph{Predicting RMPs finetuned} agent and \emph{Predicting controls finetuned} agent reached the goal successfully. Surprisingly, finetuning does not help with the \emph{Predicting depth} agent, because it is difficult to synchronize laser scans and visual images without special hardware mechanism. This results in misalignment between the predicted depth and the ground truth depth, producing spurious high loss that makes it difficult for the network to adapt to real images.

\subsubsection{Obstacle avoidance}
\label{sec:real_obstacle_avoidance}
Our hallway experiment does not show a noticeable difference between predicting RMPs and predicting controls. This is probably due to the hallway environment being simple and obstacle-free. To test them in a more challenging scenario,  we disabled the planner and put the goal point in a different hallway that is invisible from the vehicle. Furthermore, we placed two obstacles that were not seen during training as shown in Fig.~\ref{fig:real_obstacle_avoidance}a. 

When the initial heading of the vehicle was pointing in the free space direction, both \emph{Predicting RMPs} and \emph{Predicting controls} were able to reach the goal while avoiding obstacles (Fig.~\ref{fig:real_obstacle_avoidance}b). However, if we set the initial heading to point towards the wall (Fig.~\ref{fig:real_obstacle_avoidance}c), the \emph{Predicting controls} agent ran straightly towards the goal and failed. In contrast, our RMP agent backed the vehicle first so that it would have sufficient headroom to do a right turn, and successfully reached the goal. The RMP agent learned that there was high cost when moving forward from the expert RMP, and hence the generated RMPs performed a conservative maneuver.

\section{CONCLUSIONS}
We present a novel neural autonomous navigation framework that generates smooth obstacle avoidance behaviors while being more generalizable than directly predicting depth and control commands. The key of our approach is to utilize the RMPs to unify the geometry and dynamics of the vehicle and its interaction with the environment. Since our method models the geometry and dynamics explicitly, we believe it has strong potentials in image-based robot manipulation, policy transfer and agile robot maneuver. Future works include applying it into diverse robotic tasks, developing better neural architectures and unsupervised RMP learning.

While our RMP controller exhibits good properties, it is fundamentally a local policy and hence may get stuck in some situations (e.g., when the waypoint is behind a large concave obstacle). Thus it is more suitable to use a neural planner \cite{gupta2017cognitive} to provide high-level waypoints, and relies on the neural RMP for handling vehicle dynamics and local reactive obstacle avoidance.

\section{ACKNOWLEDGEMENTS}
This research was funded by the Honda Curious Minded Sponsored Research Agreement. We thank Patrick Lancaster for his help working with the RC car.

\bibliographystyle{IEEEtran}
\bibliography{IEEEabrv,references.bib}

\end{document}